\pdfoutput=1

\documentclass[11pt]{article}

\usepackage[]{latex/acl}

\usepackage{times}
\usepackage{latexsym}

\usepackage[T1]{fontenc}

\usepackage[utf8]{inputenc}

\usepackage{microtype}

\usepackage{inconsolata}

\usepackage[normalem]{ulem}
\usepackage{amsmath}
\useunder{\uline}{\ulined}{}%
\DeclareUrlCommand{\bulurl}{}
\usepackage{booktabs}
\usepackage{graphicx} 
\usepackage{multirow} 
\usepackage{caption}

\title{SCANNER: Knowledge-Enhanced Approach for Robust Multi-modal Named Entity Recognition of Unseen Entities}

\author{Hyunjong Ok$^1$$^{,3}$\thanks{\hspace{1mm} Work done as an intern at NAVER Cloud.}, Taeho Kil$^2$\thanks{\hspace{1mm} To be corresponded with.}, Sukmin Seo$^2$, Jaeho Lee$^1$\\
$^1$ POSTECH 
$^2$ NAVER Cloud 
$^3$ HJ AILAB \\
\texttt{hyunjong.ok@gmail.com, taeho.kil@navercorp.com}
 }
 
\begin{document}
\maketitle

\begin{abstract}
Recent advances in named entity recognition (NER) have pushed the boundary of the task to incorporate visual signals, leading to many variants, including multi-modal NER (MNER) or grounded MNER (GMNER). 
A key challenge to these tasks is that the model should be able to generalize to the entities unseen during the training, and should be able to handle the training samples with noisy annotations.
To address this obstacle, we propose SCANNER (\underline{S}pan \underline{CAN}didate detection and recognition for \underline{NER}), a model capable of effectively handling all three NER variants.
SCANNER is a two-stage structure; we extract entity candidates in the first stage and use it as a query to get knowledge, effectively pulling knowledge from various sources.
We can boost our performance by utilizing this entity-centric extracted knowledge to address unseen entities. 
Furthermore, to tackle the challenges arising from noisy annotations in NER datasets, we introduce a novel self-distillation method, enhancing the robustness and accuracy of our model in processing training data with inherent uncertainties.
Our approach demonstrates competitive performance on the NER benchmark and surpasses existing methods on both MNER and GMNER benchmarks.
Further analysis shows that the proposed distillation and knowledge utilization methods improve the performance of our model on various benchmarks.
\end{abstract}

\section{Introduction}
\label{sec:intro}

Named entity recognition (NER) is a fundamental task in natural language processing to identify textual spans that correspond to named entities in the given text, and classify them into pre-defined categories, such as persons, locations, and organizations~\citep{nersurvey}. The extracted information can be utilized for various downstream tasks, including entity linking and relation extraction.

\begin{figure}[t]
    \centering
    \includegraphics[width=1.0\columnwidth]{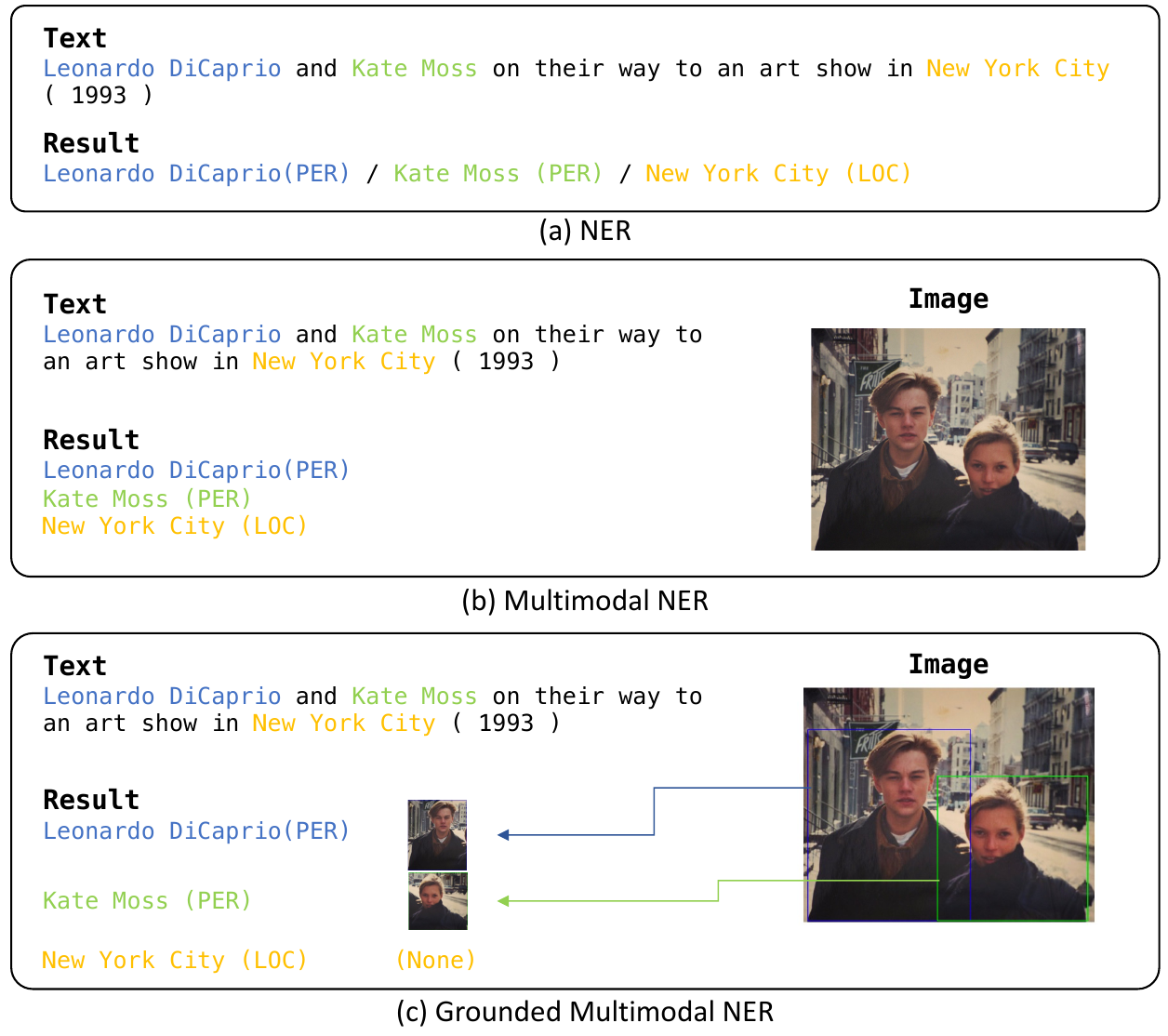}
    \caption{Illustrations of NER, MNER, and GMNER tasks. The NER task aims to identify named entities from the given text. MNER extends this task to utilize additional image informations. GMNER additionally requires the model to predict entity bounding boxes in the given image, if they are present.}
    \label{fig:taskoverview}
\end{figure}

The rapid growth of the amount of multi-modal contents on social media platforms has given rise to the multi-modal variants of NER. The most prominent example is multi-modal NER (MNER; \citet{10.5555/3504035.3504731}) , which extends traditional NER to identifying named entities in the text based on additional image input paired with the text (Fig.~\ref{fig:taskoverview}b). Another recent example is the grounded MNER (GMNER; \citet{yuetal2023grounded}); here, one additionally aims to predict the bounding boxes of named entities appearing in the given image (Fig.~\ref{fig:taskoverview}c). 

\begin{table}[t]
\centering
\resizebox{1.0\linewidth}{!}{
\begin{tabular}{llcc}
\toprule
\multirow{1}{*}{Datasets} & \multicolumn{1}{c}{Methods}
& \multicolumn{1}{c}{Seen entities}  & \multicolumn{1}{c}{Unseen entities} \\
\midrule
CoNLL2003 & BERT-base & 93.78 & 80.90 \\
& Ours (w/o Knowledge) & 96.29 & 89.68 \\
\midrule
Twitter-2015 & BERT-base & 79.81 & 57.81 \\
& Ours (w/o Knowledge) & 87.18 & 73.84  \\
\midrule
Twitter-2017 & BERT-base & 93.81 & 67.76 \\
& Ours (w/o Knowledge) & 95.68 & 82.96 \\
\bottomrule
\end{tabular}
}
\caption{A comparison of test F1 scores for the named entities that have appeared at least once in the training dataset, versus the entities that have not appeared.} 
\label{tab:ablation study non-seen entity_v1}

\end{table} 

\begin{figure}[t]
    \centering
    \includegraphics[width=1.0\columnwidth]{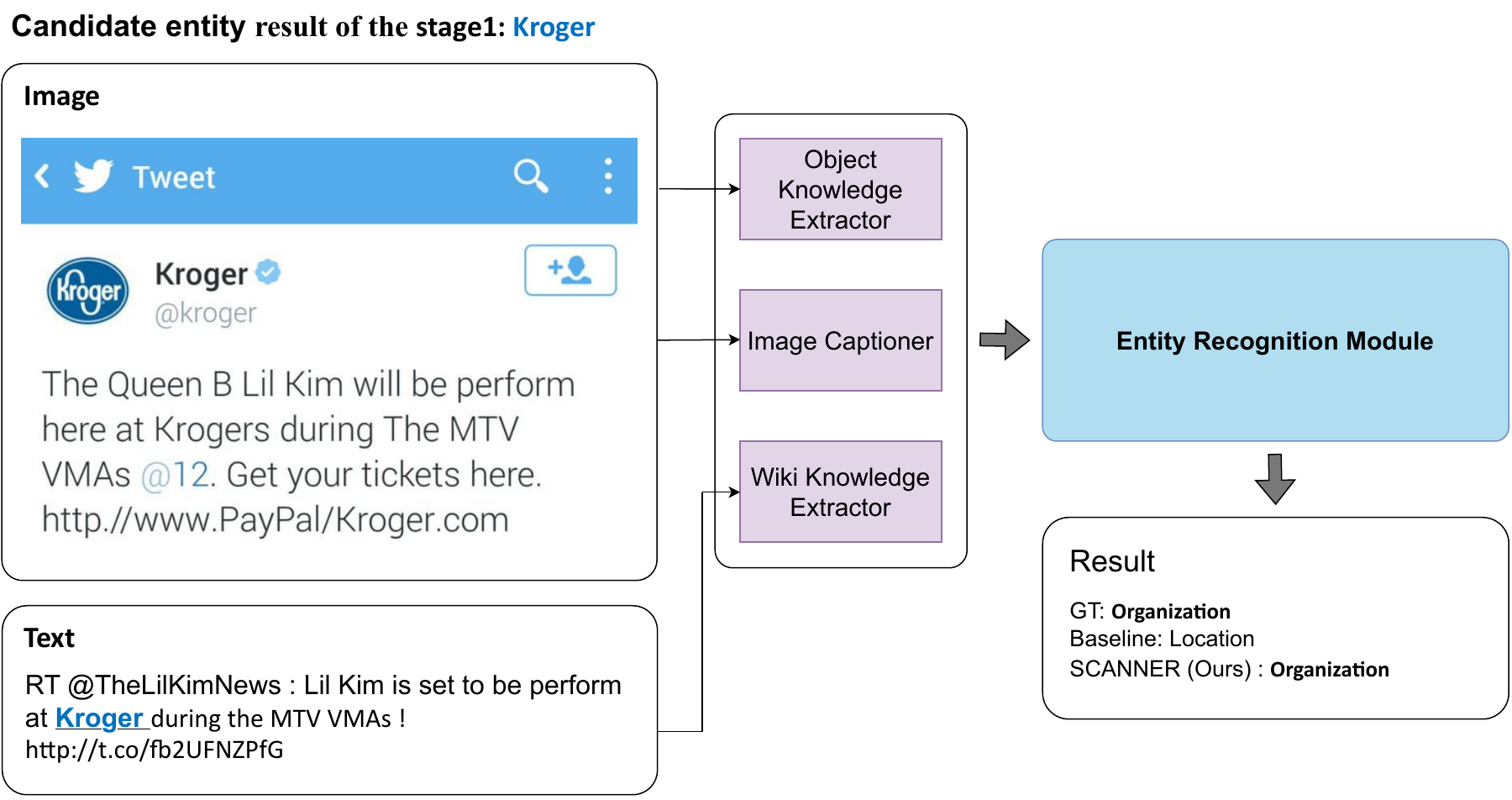}
    
    \caption{`Kroger' is an unseen entity that is hard to recognize as an Organization or Location. By our knowledge base model, it brings to successful prediction. }
    \label{fig:example_knowledgs}
\end{figure}

A major challenge in NER, MNER, and GMNER tasks is the presence of unseen entities in the test datasets, which are not found in the training datasets. Traditional models often struggle with low performance on these unseen entities (see Table~\ref{tab:ablation study non-seen entity_v1}). To tackle this problem effectively, it is important to use knowledge about unseen entities in a way that boosts ability of the model to generalize and perform well across different types of data.
In this paper, we introduce SCANNER, which stands for Span CANdidate detection and recognition for Named Entity Recognition. Our approach is designed to effectively use knowledge about unseen entities, addressing NER, MNER, and GMNER tasks with improved robustness. SCANNER adopts a two-stage structure, comprising a span candidate detection module and entity recognition module. 
The span candidate detection module identifies named entity candidates within sentences. Following this, the entity recognition module uses these candidates as queries to extract relevant knowledge from various sources, effectively recognizing the class of the entity candidate. 
As illustrated in Fig.~\ref{fig:example_knowledgs}, we were able to accurately identify `Kroger' as an `organization' by utilizing object knowledge. SCANNER effectively gathers and uses knowledge from various sources, boosting its performance in the challenging 
NER, MNER, and GMNER benchmarks. Notably, the GMNER challenge involves the intricate process of identifying entities and determining their bounding boxes within images. The architecture of SCANNER, leveraging its comprehensive knowledge, is effective in addressing the GMNER task. The effectiveness of SCANNER in the GMNER task is highlighted by establishing a new baseline that is over 21\% higher than the previous standard, as measured by the F1 score.
Additionally, we introduce the novel self-distillation method, called as Trust Your Teacher. The NER task faces challenges with noisy annotations~\citep{wang2019crossweigh, zhuli2022boundary}, particularly at entity boundaries where exact span matching is crucial and ambiguity often leads to increased noise (see Table~\ref{tab:conll2003-examples}).
Our distillation method, which softly utilizes both the prediction of the teacher model and ground truth (GT) logit, addresses the challenges of noisy annotations.

\begin{table}[t]
    \centering \small
    \resizebox{0.8\linewidth}{!}{
    \begin{tabular}{ll}
        \toprule
        Text & Dataset \\
        \midrule
        \textcolor{red}{[}The \textcolor{blue}{[}Oval\textcolor{blue}{]$_\texttt{ORG}$}\textcolor{red}{]$_\texttt{ORG}$} & CoNLL2003 \\
        \midrule
        \textcolor{blue}{[}The \textcolor{red}{[}World Cup\textcolor{red}{]$_\texttt{MISC}$}\textcolor{blue}{]$_\texttt{MISC}$} & Twitter-2015 \\
        \midrule
        \textcolor{blue}{[}Taste of \textcolor{red}{[}Toronto\textcolor{red}{]$_\texttt{LOC}$}\textcolor{blue}{]$_\texttt{MISC}$} & Twitter-2015 \\
        \midrule
        \textcolor{red}{[}Mrs. \textcolor{blue}{[}Brozik\textcolor{blue}{]$_\texttt{PER}$}\textcolor{red}{]$_\texttt{PER}$} & Twitter-2017 \\
        \midrule
        \textcolor{blue}{[}\textcolor{red}{[}Robert Downey \textcolor{red}{]$_\texttt{PER}$} Jr\textcolor{blue}{]$_\texttt{PER}$} & Twitter-2017 \\
        \bottomrule
    \end{tabular}
    }
    \caption{Examples of gold annotation and potential alternatives. The gold annotations are marked in \textcolor{blue}{blue [*]}, whereas the alternative annotations are in \textcolor{red}{red [*]}.}
    \label{tab:conll2003-examples}
\end{table}

Our approach demonstrates competitive performance on NER  and surpasses existing methods on both MNER and GMNER.
Further analysis shows that the proposed distillation and knowledge utilization methods improve the performance of our model on various benchmarks.

The contributions of SCANNER are summarized in three key aspects:
\begin{itemize}
\item We propose a new distillation method that softly blends the predictions of the teacher model with ground truth annotations to enhance data quality and model training.

\item We develop SCANNER, a two-stage structured model that effectively utilizes knowledge to improve performance, particularly in recognizing unseen entities.

\item The SCANNER model shows competitive performance in NER benchmarks and demonstrates higher performance than existing methods in MNER and GMNER benchmarks.
\end{itemize}

\section{Related work}

Prior works on MNER typically operates by first extracting the NER-related features from the image, and then combining these features with text features to recognize name entities. Roughly, existing works fall into two categories according to how they extract image features.

\noindent\textbf{Textual features.} Several works extract the textual metadata from the given image and utilize them as features for the subsequent NER task~\citep{wang2022ita, wangetal2022named, li2023prompt}. For instance, ITA~\citep{wang2022ita} extracts object tags, image captions, and OCR results from the given image. Similarly, \citet{li2023prompt} also extracts image captions, but additionally utilizes large language model as an implicit knowledge source to further refine the features. MoRe~\citep{wangetal2022named} takes a slightly different approach, using an image-based retrieval system to retrieve textual descriptions of the closest images in the database.

\noindent\textbf{Visual encoders.} Another line of work attempts to extract the image features using a visual encoder, such as pre-trained ResNets, ViTs, or CLIP vision encoder~\citep{wang2022cat, zhang2023reducing,chen2023learning}. The extracted features are then combined with the text features extracted from a separate text encoder, which often involves additional alignment via cross-modal attention~\citep{chen2022hybrid, lu2022flat, wang2022cat, zhang2023reducing, chen2023learning}. Notably, PromptMNER~\citep{wang2022promptmner} calculates the similarity between visual features and various text prompts to extract visual cues that are loosely related to the input text.

In this paper, we take a different path and extract the image features \textit{conditioned} on the information extracted from the given text. Up to our knowledge, even though it has related works in NER ~\citep{wang-etal-2021improving, wang-etal-2022damo,tan-etal-2023damo}, it is the first such attempt in the context of MNER, which is a more challenging task.

In addition, a new task has been introduced, which not only incorporates image inputs but also actively addresses the task of grounding entity locations within images~\citep{yuetal2023grounded}.

\section{Method}
\label{sec:method}

In this section, we first introduce the architecture of the proposed method, which comprises the span candidate detection module and the named entity recognition module (Sec.~\ref{subsec:arch}).
Then, we describe the named entity recognition module, which performs entity recognition and visual grounding in the image for each entity candidate (Sec.~\ref{subsec:entity_recognition}).
Finally, we explain a novel distillation method, named Trust Your Teacher, which is designed to robustly train our model even in the presence of noisy dataset annotations (Sec.~\ref{subsec:TYT}).

\subsection{SCANNER Architecture}
\label{subsec:arch}
\begin{figure}[t]
    \centering
    \includegraphics[width=0.9\columnwidth]{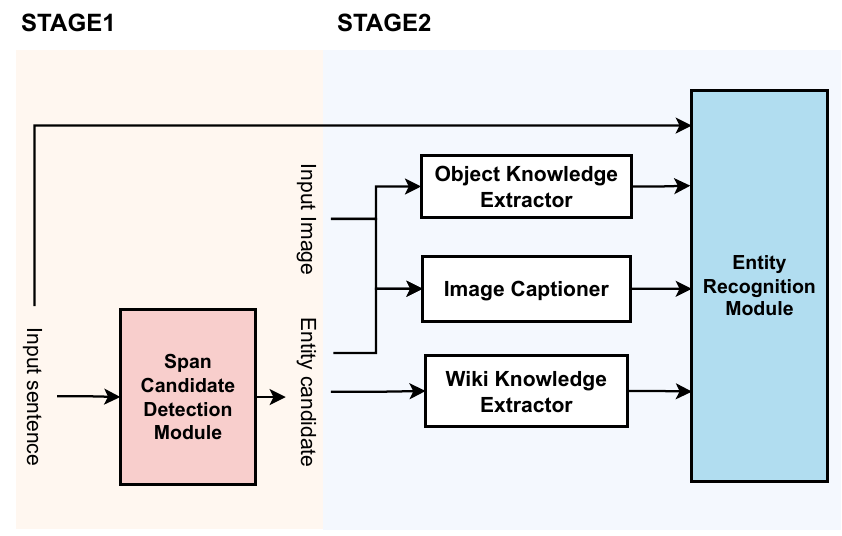}
    \caption{The overall architecture of the proposed SCANNER method. The two-stage structure allows for efficient extraction and utilization of knowledge, as knowledge is extracted only for those entity candidates that were filtered through in stage 1.}
    \label{fig:architecture}
\end{figure}

\label{subsec:stage2}
\begin{figure*}[t]
    \centering
    \includegraphics[width=2 \columnwidth]{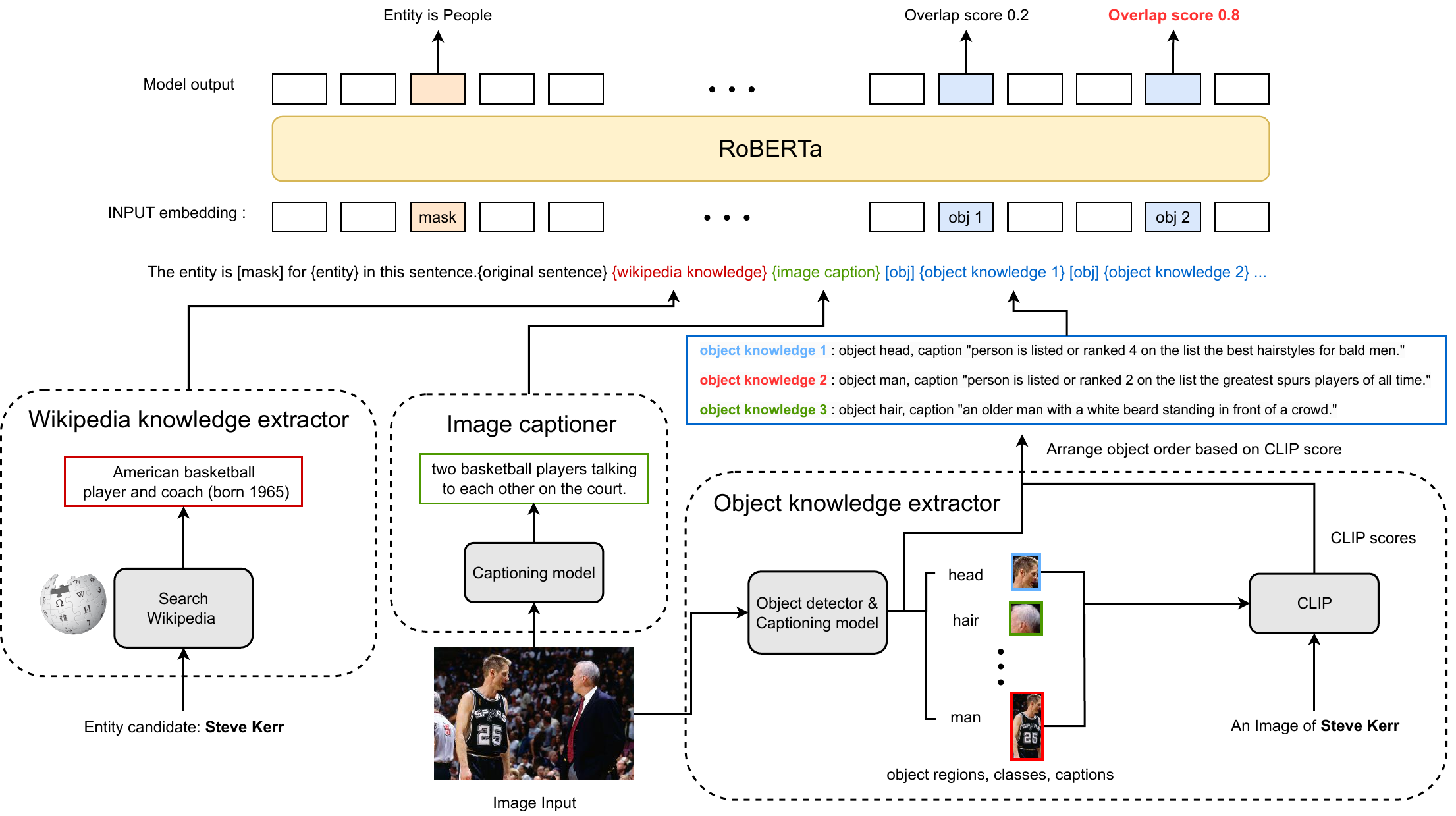}
    \caption{An illustration of the entity recognition module (stage 2). 
    Based on the entity candidates (extracted in stage 1), SCANNER utilizes various knowledge sources such as Wikipedia, image captioner, and object knowledge extractor. The knowledge collected from these sources are then processed by RoBERTa to give the final prediction.
    }
    
    \label{fig:stage2}
\end{figure*}

The primary focus of this paper is to perform MNER using both knowledge extracted from within images and external knowledge, even for entities not encountered during training.
To achieve this, as illustrated in Fig.~\ref{fig:architecture}, we propose a two-stage architecture, known for its efficiency in extracting and searching for knowledge from various sources.
In the first stage, we extract named entity candidates, and in the second stage, we efficiently search and extract only knowledge relevant to these candidates.
This acquired knowledge is then utilized for entity recognition.

\noindent\textbf{Stage 1: Span Candidate Detection Module.}
In the first stage of SCANNER, the transformer encoder~\citep{liu2019roberta} is employed to detect entity candidates from the input text.
During this phase, we utilizes BIO (Beginning, Inside, Outside) tagging to classify each token in the input text, determining whether it corresponds to the beginning, inside, or outside of an entity span.
The classification process is guided by cross-entropy loss.

\noindent\textbf{Stage 2: Entity Recognition Module.}
In Stage 2, SCANNER performs named entity recognition and visual grounding for each entity candidate detected in Stage 1.
It utilizes each entity candidate as a query to extract and leverage the necessary knowledge for the tasks.
During this process, SCANNER efficiently searches and extracts knowledge by focusing on the initially detected entity candidates rather than the entire input text.
SCANNER utilizes both internal (image-based) and external (e.g., Wikipedia) knowledge sources to perform MNER on unseen entities, not encountered in training.
Detailed information about these modules will be provided in Section~\ref{subsec:entity_recognition}.

\subsection{Entity Recognition Module} 
\label{subsec:entity_recognition}

For each entity candidate identified by the span candidate detection module, the entity recognition module processes a text prompt that includes both the entity candidate and associated knowledge.
This knowledge, extracted from images and external knowledge sources, allows for performing MNER on unseen entities that were not encountered during training.
Our methodology involves extracting this knowledge from a variety of sources, utilizing the identified entity candidates as the basis for the extraction process.
Then, this module classifies the class of each entity candidate and performs grounding to determine which object in the image corresponds to the entity.
A detailed illustration is shown in Fig.~\ref{fig:stage2}.

\subsubsection{Prompt construction with knowledge}
The entity recognition module extracts and utilizes useful knowledge from various sources when constructing the text prompt corresponding to the input.
The knowledge applied for constructing text prompts in our method includes the following.

\noindent\textbf{Wikipedia knowledge.}
Initially, information is searched using the entity candidate as a query in external knowledge source, which is Wikipedia.
This information can be valuable for classifying the type of entity for each candidate and, moreover, enables the model to classify unseen entities that were not encountered during training.
As illustrated in Fig.~\ref{fig:stage2}, for entity candidates like `Steve Kerr', it enhances entity recognition performance by providing valuable information for classification as an American basketball player and coach.

\noindent\textbf{Image caption.}
To effectively utilize visual information, image captioning results are also used.
We use the BLIP-2~\citep{li2023blip} to extract synthetic captions for the whole image.

\noindent\textbf{Object knowledge.}
In addition to global information about the image, object-level information is also beneficial for entity recognition.
To achieve this, results obtained from the object detector are employed as knowledge.
Initially, object classes are converted into text format and used as knowledge.
Then, synthetic captions for each object region are also utilized in conjunction with class names.
This information is structured as details corresponding to each object, along with a special token denoted as $[$obj$]$, as shown in Fig.~\ref{fig:stage2}.
Additionally, during this process, the visual-language similarity between each object and entity candidate is calculated, and objects are arranged in order of high similarity, which is then included in the text prompt.
One of the problems with existing methods for the MNER task is that the model sometimes references objects in the image that are irrelevant to the entity, leading to incorrect recognition.
By arranging the object details in the text prompt according to the visual-language similarity order with the entity, our model can focus more on the object regions that are highly related to the entity.
In this paper, CLIP~\citep{radford2021learning} is employed for visual-language similarity, specifically calculating the similarity between the text representation of the entity candidate and the visual representation of each Region of Interest (RoI).

All such knowledge mentioned above is converted into a textual format and integrated with the text prompt for entity recognition and visual grounding.

The text prompt, structured to include entity candidates, the entire input text sentence, and extracted knowledge, is presented as \textit{"The entity is $[$mask$]$ for \{entity\} in this sentence. \{original sentence\} \{Wikipedia\} \{image caption\} $[$obj$]$ \{object 1\} $[$obj$]$ \{object 2\} ..."}.

\subsubsection{Encoder and Objective}
The prompts constructed for each entity candidate are input into a transformer encoder model~\citep{liu2019roberta}.
For entity recognition, the output token representation of the $[$mask$]$ token in the text prompt $x_{i}$ for the $i$-th entity candidate is fed into a linear layer to predict the probability distribution $\hat{y}_{i}$.
Given the ground truth $y$, the objective function is to minimize the cross-entropy loss between the predicted entity class distribution and the ground truth logit:
\begin{equation}
\label{eq:recog_loss}
\mathcal{L}_{c} = - \sum_{i=1}^{N} y_{i} \log \hat{y}_{i},
\end{equation}
where $N$ is the total number of the entity candidates.

Additionally, the visual grounding is performed by feeding the output token representation of the $j$-th $[$obj$]$ token from the text prompt $x_{i}$ into a linear layer.
This is followed by a sigmoid function, which aids in predicting the overlap score $\hat{o}_{ij}$ between the ground truth image region grounding entity candidate $i$ and object $j$.
The objective function of visual grounding is calculated based on the binary cross-entropy loss between the overlap score and the ground truth Intersection over Union (IoU):
\begin{equation}
\begin{aligned}
\mathcal{L}_{g} = - & \sum_{i=1}^{N} \sum_{j} o_{ij} \log \hat{o}_{ij}\\
& + (1 - {o}_{ij}) \log (1 - \hat{o}_{ij}),
\end{aligned}
\end{equation}
where ${o}_{ij} $ is the ground truth IoU between the ground truth image region of the entity $i$ and object region $j$.

In training stage, we combine two losses as the final loss of our model:
\begin{equation}
\mathcal{L} = \mathcal{L}_{c} + \lambda \mathcal{L}_{g},
\end{equation}
where $\lambda$ is the weighting coefficient, we set $\lambda$ to 1 for the GMNER task and to 0 for the NER and MNER tasks in this paper.

\subsection{Trust Your Teacher}
\label{subsec:TYT}

We introduce the novel self-distillation method, called as Trust Your Teacher (TYT).
Our distillation method, which softly utilizes both the prediction of the teacher model and ground truth (GT) logit, addresses the challenges of noisy annotations.
First, we train the teacher model using equation~\ref{eq:recog_loss}, and then train the final student model using both the predictions of the teacher model and the ground truth labels.
The most significant feature of our proposed method is that it assesses the reliability of each sample by utilizing the prediction of the teacher model to determine if it is trustworthy or noisy.
Based on this assessment, the method sets the weights between the model prediction and the gt label, which are then reflected in the loss calculation.
The objective of the our proposed distillation method composes a cross-entropy loss with ground truth and Kullback-Leibler Divergence (KLD) loss with teacher predictions:
\begin{equation}
\begin{aligned}
& \mathcal{L}_{\mathrm{TYT}} = \sum_{i} a_{i} \, \mathcal{L}_{\mathrm{CE}} \big( {y}_{i} , S(x_{i}, \theta_{S}) \big) \\
& + (1 - a_{i}) \mathcal{L}_{\mathrm{KLD}} \big( S(x_{i}, \theta_{S}), T(x_{i}, \theta_{T}) \big),
\end{aligned}
\end{equation}
where $x_{i}$ is the input sample, $\theta_{S}$ and $\theta_{T}$ are the model parameters of the student and teacher, $S$ and $T$ are the prediction distributions of the student and teacher and $a_i$ is a balancing factor proposed in this paper.
In detail, $a_i$ determines whether to trust the teacher model prediction or the ground truth, and it represents the prediction score of the teacher model for the ground truth class index, which is $T(x_{i}, \theta_{T})[{y}_i]$.
This implies that since the teacher model is well-trained, if the score for the ground truth class is high, then the sample is considered reliable and more weight is given to the cross-entropy with the ground truth label.
Conversely, if the score is low, the sample is assumed to be an unreliable, noisy sample, and more weight is placed on the KLD loss with the prediction of the teacher model, rather than the ground truth label.

\begin{figure}[t]
    \centering
    \includegraphics[width=1.05\columnwidth]{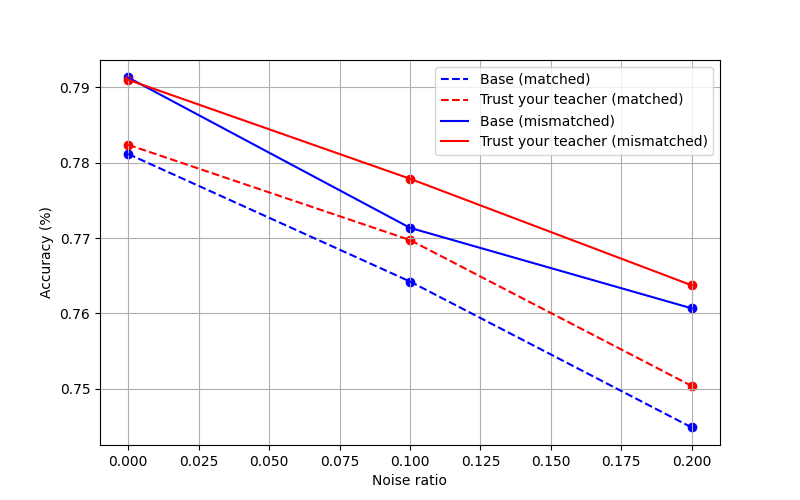}
    \caption{Experiments of text classification task in MNLI datasets. `matched' is in-domain, and `mismatched` is out-domain.} \label{fig:accuracy_noise_ratio}
\end{figure}

\begin{table}[t]
\centering
\vspace{0.25cm}
\resizebox{1\linewidth}{!}{%
\begin{tabular}{lccccccccc}
\toprule
\multirow{2}{*}{Methods} & \multicolumn{3}{|c|}{\textbf{Twitter-2015}}  &\multicolumn{3}{|c|}{\textbf{Twitter-2017}} &\multicolumn{3}{c}{\textbf{Twitter-GMNER}} \\ 
& \multicolumn{1}{|c}{Pre.}  & Rec.   & F1   & \multicolumn{1}{|c}{Pre.} &Rec. &F1 & \multicolumn{1}{|c}{Pre.} &Rec. &F1 \\ \hline
Base & 83.28 & 87.68 & 85.43 & \textbf{90.96}  &	93.23 &	92.08 & 86.60 & 87.59 & 87.09 \\
Half & 83.36 & 87.69 & 85.47 & 90.31 & 92.92 & 91.60 & \textbf{87.00} & 87.96 & 87.47\\
Full & \textbf{83.63} & 87.72 & 85.63 & 90.53 & 92.95 & 91.72 & 86.90 & 87.80 & 87.35 \\
TYT & 83.59 & \textbf{87.98} &\textbf{85.73} & 90.94 & \textbf{93.28} & \textbf{92.09} & 86.82 & \textbf{88.37} & \textbf{87.59} \\
\hline
\end{tabular}
}
\caption{Ablation study on the MNER dataset in first stage. `Half' is when $a_{i}$ is 0.5 and `Full' is 0. In `TYT', $a_{i}$ is adjusted through the trust your teacher method.
} 
\label{tab:ablation study stage1}
\end{table} 

To demonstrate the significant impact of our TYT approach, we have carried out some experiments.
Fig.~\ref{fig:accuracy_noise_ratio} illustrates our experiments on a text classification task in MNLI dataset.
We extract about 30\% of the train set for experimental efficiency and intentionally added label noise at rates of 10\% and 20\% to this subset.
We then compare the performance of the model trained with our TYT method on the train set with added label noise against the baseline that does not use distillation.
Fig.~\ref{fig:accuracy_noise_ratio} indicates that using TYT demonstrates relatively robust performance under moderate noise conditions.
Additionally, we compare our method with the conventional soft distillation methods that do not dynamically vary the $a_{i}$ parameter in the entity detection task, stage 1 of MNER and GMNER.
Table~\ref{tab:ablation study stage1} shows that our method has better performance on MNER and GMNER benchmarks, and adaptively varying the $a_{i}$ is more effective than keeping it fixed.

We apply the TYT to both stages 1 and 2. But in NER, we only use it in stage 1.
The loss from the TYT is applied only to the classification loss and not to the loss for visual grounding.

\section{Experiment}

\subsection{Dataset}
Our methodology's efficacy was assessed using widely used datasets for each task. 
We utilize CoNLL2003~\citep{tjong-kim-sang-de-meulder-2003} for NER, Twitter-2015~\citep{10.5555/3504035.3504731} and Twitter-2017~\citep{luetal2018visual} for MNER, and Twitter-GMNER~\citep{yuetal2023grounded} for GMNER. Details are in appendix~\ref{detail data information}.

\begin{table*}[!]
\small
\setlength\tabcolsep{1.7pt}
\renewcommand{\arraystretch}{1.0}
\centering
\resizebox{\textwidth}{!}{
\begin{tabular}{lcccccccccccccc}
\toprule
 \multirow{3}{*}{Methods} 
 & \multicolumn{7}{|c|}{\textbf{Twitter-2015}}  &\multicolumn{7}{c}{\textbf{Twitter-2017}}  \\\cline{2-15}
 & \multicolumn{4}{|c|}{\textbf{Single Type(F1)}} & \multicolumn{3}{c|}{\textbf{Overall}} & \multicolumn{4}{c|}{\textbf{Single Type(F1)}} & \multicolumn{3}{c}{\textbf{Overall}}\\
  &\multicolumn{1}{|c}{PER}   & LOC   & ORG  & \multicolumn{1}{c|}{OTH.}  & Pre.   & Rec.   & \multicolumn{1}{c|}{F1} & PER   & LOC   & ORG   & \multicolumn{1}{c|}{OTH.}  & Pre. &Rec. &F1   \\ 
\midrule 
 \multicolumn{15}{c}{ Text}\\
\midrule
BERT-CRF$^\dag$ & \multicolumn{1}{|c}{85.37} & 81.82 & 63.26 & \multicolumn{1}{c|}{44.13} & 75.56 & 73.88 & \multicolumn{1}{c|}{74.71} & 90.66 & 84.89 & 83.71 & \multicolumn{1}{c|}{66.86} & 86.10 & 83.85 & 84.96  \\
BERT-SPAN$^\dag$ \cite{yamada-etal-luke} & \multicolumn{1}{|c}{85.35} & 81.88 & 62.06 & \multicolumn{1}{c|}{43.23} & 75.52 & 73.83 & \multicolumn{1}{c|}{74.76} & 90.84 & 85.55 & 81.99 & \multicolumn{1}{c|}{69.77} & 85.68 & 84.60 & 85.14 \\
RoBERTa-SPAN$^\dag$ \cite{yamada-etal-luke} & \multicolumn{1}{|c}{87.20} & 83.58 & 66.33 & \multicolumn{1}{c|}{50.66} & 77.48 & 77.43 & \multicolumn{1}{c|}{77.45} & 94.27 & 86.23 & 87.22 & \multicolumn{1}{c|}{74.94} & 88.71 & 89.44 & 89.06 \\
\midrule 
 \multicolumn{15}{c}{Vision-LLM (w/ zero-shot)}\\
\midrule
Gemini-pro-vision & \multicolumn{1}{|c}{73.12} & 65.53 & 35.80 & \multicolumn{1}{c|}{20.72} & 48.24 & 64.88 & \multicolumn{1}{c|}{55.34} & 84.36 & 71.65 & 61.24 & \multicolumn{1}{c|}{22.02} & 64.02 & 69.90 & 66.83 \\
GPT4-V & \multicolumn{1}{|c}{80.00} & 75.26 & 40.53 & \multicolumn{1}{c|}{25.26} & 51.46 & 70.42 & \multicolumn{1}{c|}{59.46} & 85.63 & 78.62 & 73.68 & \multicolumn{1}{c|}{36.63} & 67.63 & 74.90 & 71.08 \\
\midrule 
\multicolumn{15}{c}{ Text+Image}\\
\midrule
UMT \cite{yu-etal-2020-improving-multimod} & \multicolumn{1}{|c}{85.24} & 81.58 &  63.03 &  \multicolumn{1}{c|}{39.45} &  71.67 &  75.23 &  \multicolumn{1}{c|}{73.41} &  91.56 &  84.73 & 82.24 &  \multicolumn{1}{c|}{70.10} &  85.28 &  85.34 & 85.31 \\
UMGF \cite{Zhang_Wei_Li_Wu_Zhu_Zhou_2021} & \multicolumn{1}{|c}{84.26} & 83.17 & 62.45 &\multicolumn{1}{c|}{42.42} & 74.49 & 75.21 &  \multicolumn{1}{c|}{74.85} &  91.92 &  85.22 &  83.13 &
  \multicolumn{1}{c|}{69.83} &  86.54 &  84.50 & 85.51 \\
MNER-QG \cite{jia2023mner} & \multicolumn{1}{|c}{85.68} &  81.42 &  63.62 &  \multicolumn{1}{c|}{41.53} &77.76 & 72.31 &  \multicolumn{1}{c|}{74.94} &  93.17 &  86.02 & 84.64 &  \multicolumn{1}{c|}{71.83} & 88.57 &  85.96 &  87.25 \\ 
R-GCN \cite{10.1145/3503161.3548228} &  \multicolumn{1}{|c}{86.36} &  82.08 &  60.78 &\multicolumn{1}{c|}{41.56} &  73.95 &  76.18 &  \multicolumn{1}{c|}{75.00} &  92.86 &  86.10 &  84.05 &  \multicolumn{1}{c|}{72.38} &  86.72 &  87.53 &  87.11 \\
ITA \citep{wang2022ita} &  \multicolumn{1}{|c}{-} &  - &  - &  \multicolumn{1}{c|}{-} &  - &  - &  \multicolumn{1}{c|}{78.03} &  - &  - &  - &  \multicolumn{1}{c|}{-} &  - &  - &  89.75 \\
PromptMNER \cite{wang2022promptmner} &  \multicolumn{1}{|c}{-} &  - &  - &  \multicolumn{1}{c|}{-} &  78.03 &  79.17 &  \multicolumn{1}{c|}{78.60} &  - &  - &  - &\multicolumn{1}{c|}{-} &  89.93 &  90.60 &  90.27 \\
CAT-MNER \cite{wang2022cat} &  \multicolumn{1}{|c}{88.04} &  84.70 &  68.04 &  \multicolumn{1}{c|}{52.33} &  78.75 &  78.69 &  \multicolumn{1}{c|}{78.72} &  94.61 &  88.40 &  88.14 &  \multicolumn{1}{c|}{80.50} &  90.27 &  90.67 &  90.47\\
MoRe \cite{wangetal2022named} &  \multicolumn{1}{|c}{-} &  - &  - &  \multicolumn{1}{c|}{-} &  - &  - &
\multicolumn{1}{c|}{79.21} &  - &- &  - &  \multicolumn{1}{c|}{-} &  - &  - &  90.67 \\
PGIM $^\ddag$ \cite{li2023prompt} &  \multicolumn{1}{|c}{88.34} &  84.22 &  70.15 &  \multicolumn{1}{c|}{52.34} &  79.21 &  \textbf{79.45} &  \multicolumn{1}{c|}{79.33} &  96.46 &  89.89 &  89.03 &  \multicolumn{1}{c|}{79.62} &  \textbf{90.86} &  \textbf{92.01} & \textbf{91.43} \\
 \textbf{SCANNER} (Ours) &  \multicolumn{1}{|c}{88.24} &  85.16 &  69.86 &  \multicolumn{1}{c|}{52.23} &  \textbf{79.72} &  79.03 &  \multicolumn{1}{c|}{\textbf{79.38}} &  
 95.18 &  88.52 &  88.45 &  \multicolumn{1}{c|}{79.71} &  90.40 & 90.67 & 90.54 \\
 & \multicolumn{1}{|c}{$\pm$ 0.27 } & {$\pm$ 0.22 } & {$\pm$ 0.31 }  & \multicolumn{1}{c|}{$\pm$ 1.39 } & {$\pm$ 0.56 } & {$\pm$ 0.64 } & \multicolumn{1}{c|}{$\pm$ 0.14 }  
 & {$\pm$ 0.23 } & {$\pm$ 0.26 } & {$\pm$ 0.66 }  & \multicolumn{1}{c|}{$\pm$ 2.98 } & {$\pm$ 0.19 } & {$\pm$ 0.53 } & {$\pm$ 0.32 }\\

\bottomrule
\end{tabular}
}
\caption{Experiment results on the Twitter-2015 and Twitter-2017. The results for methods marked with $^\dag$ are from~\citet{wang2022cat}. The methods marked with $^\ddag$ denotes that they utilize LLMs (of ChatGPT scale) as knowledge sources.}
\label{tab:experiments_result_mner}
\end{table*}
\begin{table}[t]
\centering
\vspace{0.25cm}
\resizebox{1\linewidth}{!}{
\begin{tabular}{lccc}
\toprule
\multirow{2}{*}{Methods} 
 & \multicolumn{3}{c}{\textbf{CoNLL2003}} \\
& \multicolumn{1}{c}{Pre.}  & Rec.   & F1 \\

\midrule 
W$^2$NER \cite{Li_Fei_Liu_Wu_Zhang_Teng_Ji_Li_2022} & 92.71 & \textbf{93.44} & 93.07 \\
DiffusionNER \cite{shen-etal-2023-diffusion} & 92.99 & 92.56 & 92.78 \\
PromptNER \cite{shen-etal-2023-promptne} & 92.96 & 93.18 & 93.08 \\
SCANNER (Ours) & \textbf{93.07} & \textbf{93.44} & \textbf{93.26} \\
 & {$\pm$ 0.20 } & {$\pm$ 0.23 } & {$\pm$ 0.21 } \\

\bottomrule
\end{tabular}
}
\caption{Experiment results on the CoNLL2003.}
\label{tab:experiments_result_NER}
\end{table}

\subsection{Experimental Setups}   
\noindent\textbf{Evaluation metrics.}
To evaluate our method, we use Entity-wise F1, precision, and recall scores for NER and MNER tasks. For the GMNER task, there is an additional evaluation of the visual grounding. For instances, where the visual grounding is ungroundable, a prediction is correct if it is classified as `None.' For others, correctness hinges on the IoU metric. A prediction is considered correct if the IoU score between the predicted visual region and the ground truth bounding boxes exceeds a threshold of 0.5. We use F1, precision, and recall scores, which are calculated based on the aggregate correctness across entity, type, and visual region predictions. Our primary focus is on the F1 score in line with numerous preceding studies.

\noindent\textbf{Implementation details.}
Following most recent works, we implement our model utilizing RoBERTa-large in NER, XLM-RoBERTa-large~\citep{conneau-etal-2020-unsupervise} for MNER, GMNER both in stage 1 and stage 2. For the object detector, we use VinVL~\citep{zhang2021vinvl} following the settings with ITA~\citep{wang2022ita}. To address the requirements of visual-language similarity and image caption, we use each of them CLIP~\footnote{openai/clip-vit-large-patch14} and BLIP-2~\footnote{salesforce/blip2-opt-2.7b} models respectively. Detailed hyper-parameter settings are shown in appendix~\ref{detail hyper-parameter}. All experiments were done on a single GeForce RTX 4090 GPU or NVIDIA H100 GPU, and we report the average score from 5 runs with different random seeds for each setting.

Also we applied several minor methods to enhance performance. In the second stage, we incorporated a `non-entity' label to account for instances where the model erroneously predicts entity candidates not present in the dataset. That allowed for more accurate handling of such cases. We augmented it with non-entity data by dividing the training set into four folds in stage 1 and validating each fold.
Secondly, we employed adversarial weight perturbation (AWP)~\citep{NEURIPS2020_1ef91c21} in stage 1
, which enhances the robustness and generalization capabilities of the model. 
We initiated AWP from an intermediate stage of our training process.

\subsection{Experimental results in various NER tasks}

\noindent\textbf{Experimental results in NER.}
To evaluate the effectiveness of our approach in NER, we primarily compared our model against the existing methods in Table~
\ref{tab:experiments_result_NER}. It shows that SCANNER exhibits a competitive performance compared to the existing NER methods.

\noindent\textbf{Experimental results in MNER.}
In assessing the effectiveness of SCANNER in MNER, we conducted comparative analyses against various leading models in this task. The results, detailed in Table~\ref{tab:experiments_result_mner}, reveal that our model achieves superior performance in Twitter-2015 and exhibits markedly impressive results in Twitter-2017. 
Notably, while PGIM shows outstanding performance on Twitter-2017, it utilizes large language models (LLM) like ChatGPT, which incurs API costs, a notable drawback. In contrast, our model does not rely on LLM knowledge, freeing it from such disadvantages and demonstrating better performance on Twitter-2015. Additionally, we conduct experience using the same LLM knowledge as PGIM, which is in appendix~\ref{Compare-with-MoRe-and-PGIM}.

\begin{table}[t]
\centering
\vspace{0.25cm}
\resizebox{\linewidth}{!}{
\begin{tabular}{lccc}
\toprule
\multirow{2}{*}{Methods} 
 & \multicolumn{3}{c}{\textbf{Twitter-GMNER}} \\
& \multicolumn{1}{c}{Pre.}  & Rec.   & F1 \\
\midrule 
 \multicolumn{4}{c}{ Text}\\
\midrule
HBiLSTM-CRF-None \cite{luetal2018visual}  & 43.56 & 40.69 & 42.07 \\
BERT-None \cite{devlin-etal-2019-ber}  & 42.18 & 43.76 & 42.96 \\
BERT-CRF-None & 42.73 & 44.88 & 43.78 \\
BARTNER-None \cite{yan2021unified}  & 44.61 & 45.04 & 44.82 \\
\midrule 
\multicolumn{4}{c}{ Text+Image}\\
\midrule
GVATT-RCNN-EVG \cite{luetal2018visual}  & 49.36 & 47.80 & 48.57 \\
UMT-RCNN-EVG \cite{yu-etal-2020-improving-multimod} & 49.16 & 51.48 & 50.29 \\
UMT-VinVL-EVG \cite{yu-etal-2020-improving-multimod} & 50.15 & 52.52 & 51.31 \\
UMGF-VinVL-EVG \cite{Zhang_Wei_Li_Wu_Zhu_Zhou_2021}  & 51.62 & 51.72 & 51.67 \\
ITA-VinVL-EVG  \citep{wang2022ita} & 52.37 & 50.77 & 51.56 \\
BARTMNER-VinVL-EVG \cite{yuetal2023grounded} & 52.47 & 52.43 & 52.45 \\
H-Index \cite{yuetal2023grounded} & 56.16 & 56.67 & 56.41 \\
SCANNER (Ours) & \textbf{68.34} & \textbf{68.71} & \textbf{68.52}  \\
 & {$\pm$ 0.73 } & {$\pm$ 0.61 } & {$\pm$ 0.67 } \\
\bottomrule
\end{tabular}
}
\caption{Experiment results on the Twitter-GMNER. The reported figures for the baseline models are taken from~\citet{yuetal2023grounded}.}
\label{tab:experiments_result_GMNER}
\end{table}

\begin{table*}[t]
\centering
\vspace{0.25cm}
\resizebox{\textwidth}{!}{%
\begin{tabular}{c|c}
\multicolumn{2}{c}{\includegraphics[width=1.6180339887\columnwidth]{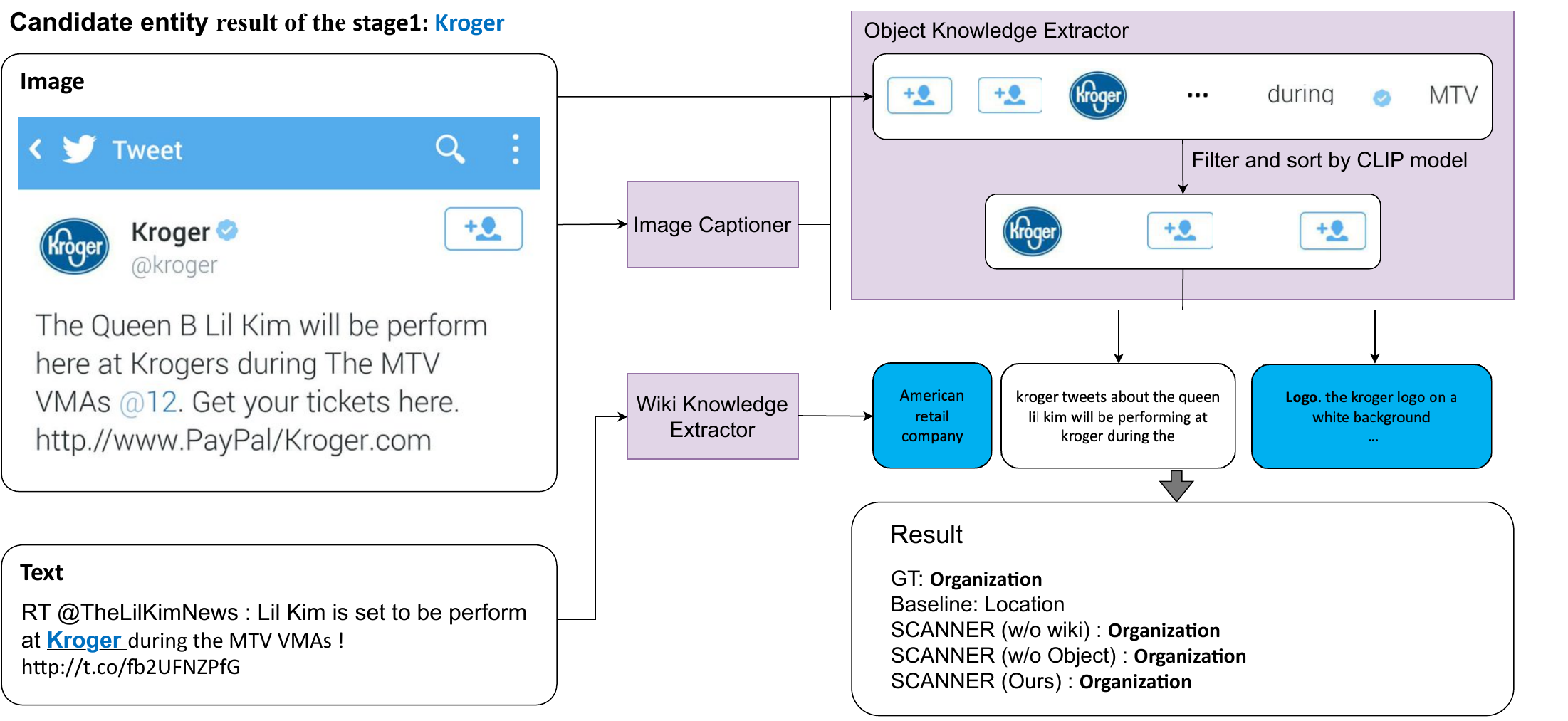} }\\
 \hline
 \noalign{\smallskip}
 \centering
    \includegraphics[width=1.0\columnwidth]{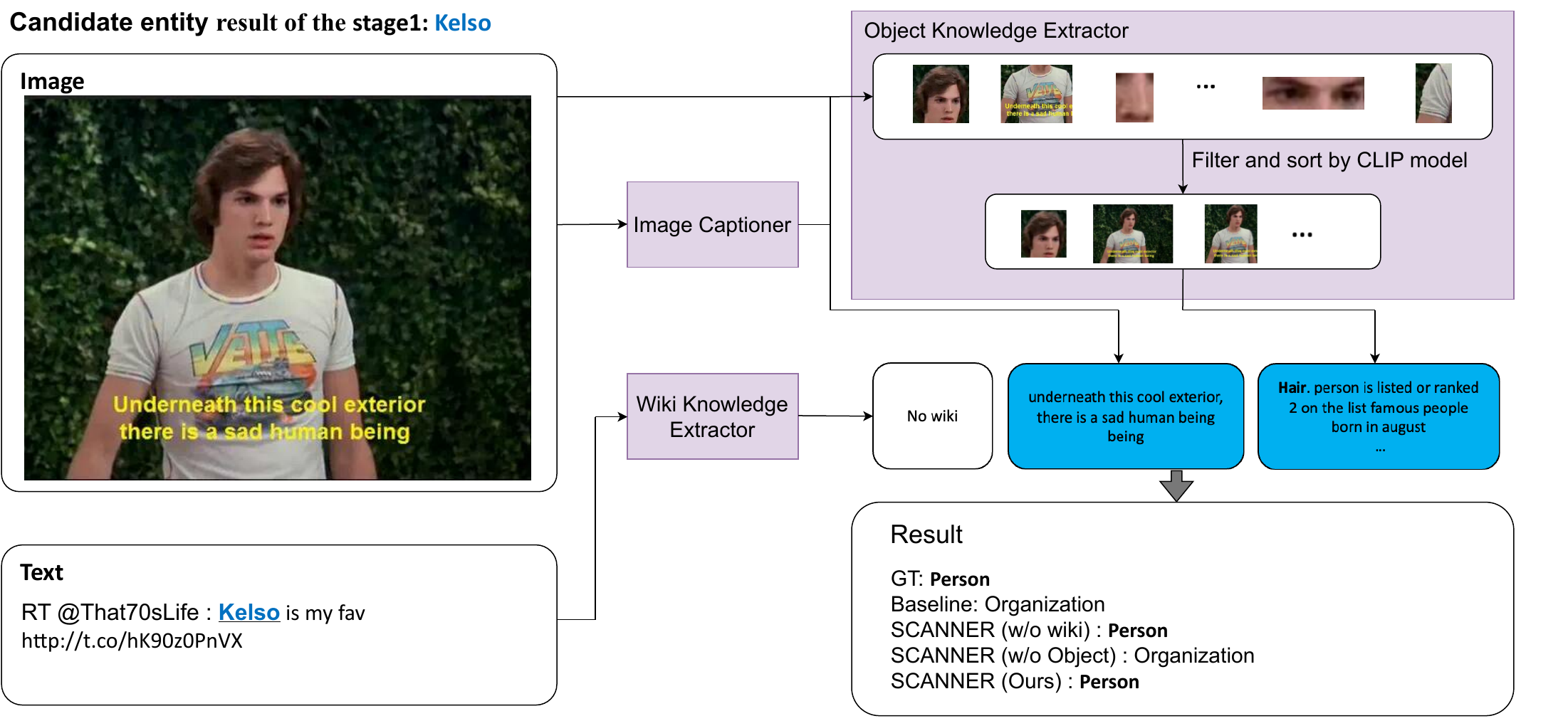} & 
    \includegraphics[width=1.0\columnwidth]{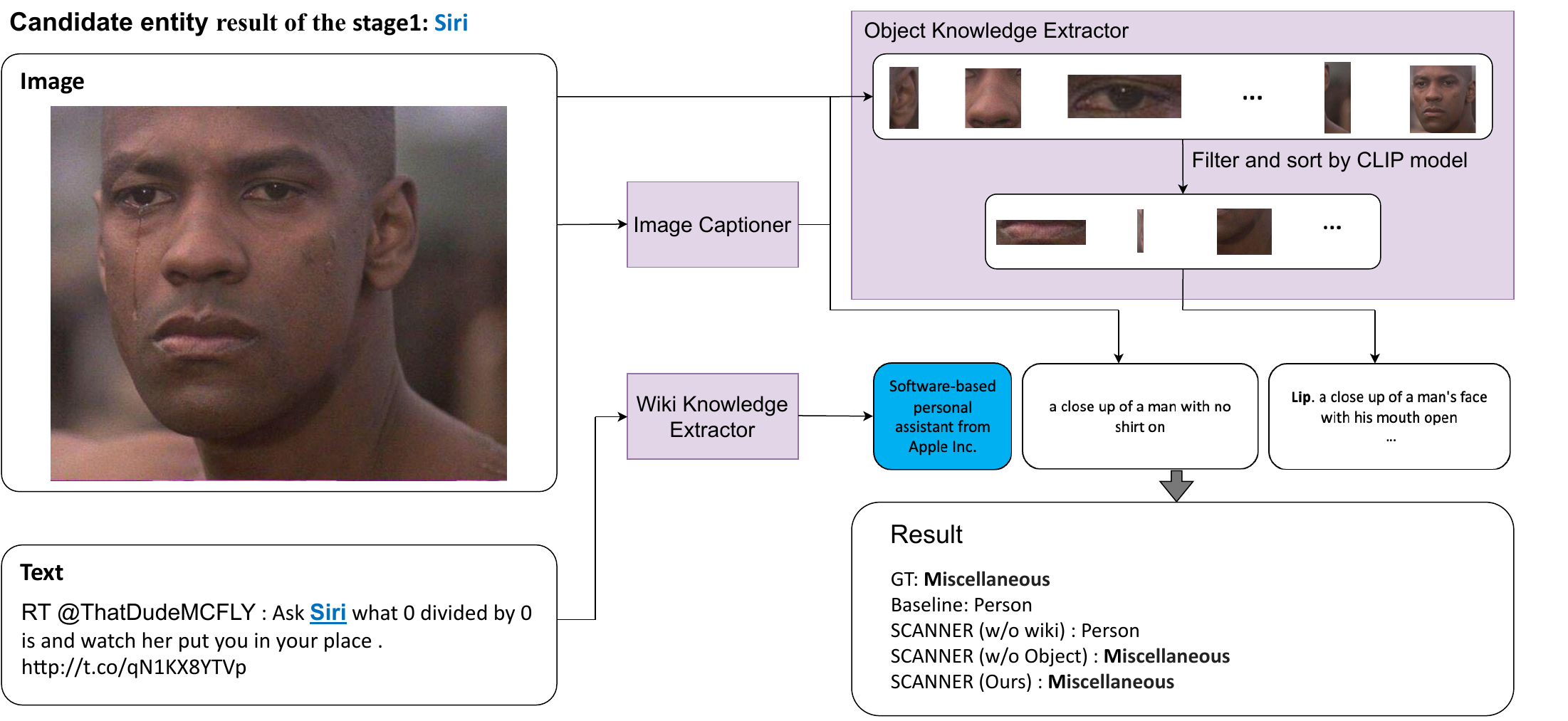}
    \\
\end{tabular}%
}
\captionof{figure}{Visualization results showing how various types of knowledge are brought in and utilized differently to perform the MNER task. Knowledge highlighted in blue positively influences correct predictions.} 
\label{tab:Case study}
\end{table*} 

\noindent\textbf{Experimental results in GMNER.} To show our effectiveness in GMNER, we make broad comparisons with all existing methods. Text-only models made to predict the visual groundings all `None'. The Table~\ref{tab:experiments_result_GMNER} shows that our model achieves significant performance improvements over prior research and establishes a new powerful baseline for future GMNER studies.

\subsection{Ablation study}

\begin{table}[t]
\centering
\vspace{0.25cm}
\resizebox{0.9\linewidth}{!}{%
\begin{tabular}{lcccccc}
\toprule
\multirow{2}{*}{Methods} & \multicolumn{3}{|c|}{\textbf{Twitter-2015}}  &\multicolumn{3}{c}{\textbf{Twitter-2017}} \\ 
& \multicolumn{1}{|c}{Pre.}  & Rec.   & F1   & \multicolumn{1}{|c}{Pre.} &Rec. &F1 \\
\midrule
SCANNER & 79.72 & 79.03 & 79.38 & 90.40 & 90.67 & 90.54\\
\quad - TYT & -0.26 & -0.17 & -0.21 & -0.24 & -0.11 & -0.18 \\
\quad - OBK & +0.11 & -0.60 & -0.26 & -0.23 & -0.22 & -0.22\\
\quad - WKK & -1.12 & -0.14 & -0.64 & -0.51 & -0.44 & -0.48\\
\quad - ICK & -0.08 & -0.54 & -0.31 & -0.29 & -0.31 & -0.29 \\
\bottomrule
\end{tabular}%
}

\caption{Ablation studies on MNER datasets. `-TYT' is without trust your teacher method. `-OBK' is without object knowledge. `-WKK' is without Wikipedia knowledge.`-ICK' is without image caption knowledge.} 
\label{tab:ablation study}
\end{table} 

\noindent\textbf{Ablation study in MNER.}
We conduct ablation experiments on the MNER task to evaluate the effectiveness of the proposed method.
These results are shown in Table~\ref{tab:ablation study}. We observe that removing the Trust Your Teacher method led to a decrease in performance. Our proposed distillation method effectively alleviates the dataset noise issue, making our model more robust to learning from noisy dataset.
Additionally, to verify the effectiveness of the various types of knowledge used in our study, we compare the results with experiments where each type of knowledge was removed. We confirm that the object knowledge, Wikipedia knowledge, and image caption knowledge used in our paper all contribute to the performance improvement of the MNER task.

\noindent\textbf{Case study.}
As shown in Fig.~\ref{tab:Case study}, all three types of knowledge can be utilized as useful information for named entity recognition. In the case of the first image, knowledge from Wikipedia such as "American retail company" and object knowledge containing the logo information of "Kroger" both help in predicting the "Kroger" entity as an organization.
For the image on the bottom left, image caption and object knowledge aided in named entity recognition.
Moreover, in the image on the bottom right, vision information like image caption and object knowledge led to incorrect entity recognition results, but it was corrected through external knowledge from Wikipedia.
Thus, the three types of knowledge proposed in this paper complement each other, enabling accurate MNER performance.

\begin{table}[t]
\centering
\vspace{0.25cm}
\resizebox{0.8\linewidth}{!}{%
\begin{tabular}{lcccc}
\toprule
\multirow{2}{*}{Datasets} & \multicolumn{2}{|c}{\textbf{w/o Knowledge}} 
& \multicolumn{2}{|c}{\textbf{w/ Knowledge} }\\ 
& \multicolumn{1}{|c}{Seen}  & \multicolumn{1}{c}{Unseen}  
& \multicolumn{1}{|c}{Seen}  & \multicolumn{1}{c}{Unseen}\\ 
\midrule
CoNLL2003 & 96.29 & 89.68 & 96.35 & 89.70 \\
Twitter-2015 & 87.18 & 73.84 & 87.50 & 75.45 \\
Twitter-2017 & 95.68 & 82.96 & 95.90 & 83.71 \\
\bottomrule
\end{tabular}%
}
\caption{The result comparing the test F1 scores in unseen entities of knowledge extracted and baseline.} 
\label{tab:ablation study non-seen entity}
\end{table} 

\noindent\textbf{Effectiveness in unseen entity.}
Table ~\ref{tab:ablation study non-seen entity} shows the effectiveness of knowledge in unseen entities. As SCANNER utilizes various knowledge in MNER, it greatly increases performance in unseen entities. In NER, lack of various knowledge causes there to be no image, which slightly improves the performance.

\section{Conclusions}

We introduce SCANNER, a novel approach for performing NER tasks by utilizing knowledge from various sources.
To efficiently fetch diverse knowledge, SCANNER employs a two-stage structure, which detects entity candidates first, and performs named entity recognition and visual grounding on these candidates.
Additionally, we propose the novel distillation method, which robustly trains the model against dataset noise, demonstrating superior performance in various NER benchmarks.
We believe that our method can be easily extended to utilize knowledge from multiple sources that were not covered in this paper.
\section*{Limitations}
In this study, we extract knowledge from various sources and utilize it to perform MNER tasks.
By leveraging several vision experts such as CLIP, and also fetching external knowledge, our method takes relatively longer inference time compared to approaches that do not use knowledge.
However, the use of vision experts and knowledge is essential for a MNER model that functions well even with unseen entities, and we efficiently extract information through a two-stage structure.

\section*{Ethics statement}
All experimental results we provide in this paper is based on publicly available datasets and open-source models.

\section*{Acknowledgments}
This work was partly supported by Artificial Intelligence Industrial Convergence Cluster Development project funded by the Ministry of Science and ICT (MSIT, Korea) \& Gwangju Metropolitan City, and the National Research Foundation of Korea (NRF) grant (RS-2023-00213710, Neural Network Optimization with Minimal Optimization Costs), and the Institute of Information \& communications Technology Planning \& Evaluation (IITP) grant funded by the Korea government(MSIT) (No.2019-0-01906, Artificial Intelligence Graduate School Program(POSTECH))

\bibliography{custom}

\newpage
\clearpage

\appendix

\section{Hyper-parameter settings}
\begin{table}[h]
\centering
\resizebox{\linewidth}{!}{
\begin{tabular}{lccccc}
\toprule
\multirow{2}{*}{Datasets} 
 & \multicolumn{4}{c}{\textbf{Stage 1}} \\
\cmidrule(r){2-5}
& epochs & batch size & lr & weight decay    \\
\midrule
CoNLL2003 & 5 & 8 & $5 \times 10^{-6}$ & 1  \\
Twitter-2015 & 10 & 4 & $1 \times 10^{-5}$ & 2  \\
Twitter-2017 & 10 & 8 & $1 \times 10^{-5}$ & 2  \\
Twitter-GMNER & 10 & 8 & $1 \times 10^{-5}$ & 2 \\
\bottomrule
\end{tabular}
}
\caption{Hyper-parameter settings in Stage 1 were used in the experiments for NER, MNER, and GMNER.}
\label{parametersetting_stage1}
\end{table}

\begin{table}[h]
\centering
\resizebox{\linewidth}{!}{
\begin{tabular}{lcccccc}
\toprule
\multirow{2}{*}{Datasets} 
 & \multicolumn{5}{c}{\textbf{Stage 2}} \\
\cmidrule(r){2-6} 
& epochs & batch size & lr & weight decay & max objects    \\
\midrule
CoNLL2003 & 20 & 8 & $3 \times 10^{-6}$ & 0.01 & - \\
Twitter-2015 & 5 & 8 & $1 \times 10^{-5}$ & 0.01  & 15  \\
Twitter-2017 & 7 & 8 & $5 \times 10^{-6}$ & 0.01 & 15   \\
Twitter-GMNER & 5 & 8 & $5 \times 10^{-6}$ & 2 & 18    \\
\bottomrule
\end{tabular}
}
\caption{Hyper-parameter settings in Stage 2 were used in the experiments for NER, MNER, and GMNER.}
\label{parametersetting_stage2}
\end{table}

\begin{table}[t]
\centering
\resizebox{\linewidth}{!}{
\begin{tabular}{lccccc}
\toprule
& \multicolumn{4}{c}{Text} & \multicolumn{1}{c}{Image Total} \\
\cmidrule(lr){2-5} \cmidrule(lr){6-6}
& \#Total & \#Train & \#Dev & \#Test  & \#Groundable Entity   \\
\midrule
CoNLL2003 & 20,744 & 17,291 & - & 3,453  & -  \\
Twitter-2015 & 8,257 & 4,000 & 1,000 & 3,257   & -  \\
Twitter-2017 & 4,819 & 3,373 & 723 & 723  & -   \\
Twitter-GMNER & 10,000 & 7,000 & 1,500 & 1,500 &  6,716   \\
\bottomrule
\end{tabular}
}
\caption{Dataset statistics of NER, MNER, and GMNER benchmarks}
\label{Dataset Statistics}
\end{table}
\begin{table}[t]
\centering
\vspace{0.25cm}
\resizebox{1\linewidth}{!}{%
\begin{tabular}{lcccccc}
\toprule
\multirow{2}{*}{Methods} & \multicolumn{3}{|c|}{\textbf{Twitter-2015}}  &\multicolumn{3}{c}{\textbf{Twitter-2017}} \\ 
& \multicolumn{1}{|c}{Pre.}  & Rec.   & F1   & \multicolumn{1}{|c}{Pre.} &Rec. &F1 \\
\midrule
MoRe \cite{wangetal2022named} & - & - & 79.21 & - & - & 90.67 \\
PGIM \cite{li2023prompt} & 	79.21 &	79.45 &	79.33 &	\textbf{90.86} &	92.01 &	\textbf{91.43} \\
SCANNER & \textbf{79.72} & 79.03 & 79.38 & 90.40 & 90.67 & 90.54\\
SCANNER (w/ GPT knowledge) & 79.24 & \textbf{80.97} & \textbf{80.10} & 90.22 & \textbf{92.23} & 91.22\\
\bottomrule
\end{tabular}%
}

\caption{Comparison SCANNER w/ GPT knowledge with previous leading baseline methods on Twitter-2015 and Twitter-2017 datasets.} 
\label{Using_gpt_knowledge}
\end{table} 
\begin{table}[t]
\centering
\vspace{0.25cm}
\resizebox{1\linewidth}{!}{
\begin{tabular}{lrr}
\toprule
Methods & Knowledge base & Sentences / Sec. \\

\midrule 
 MoRe & Wiki (Text) & 64.6 \\
 MoRe & Wiki (Image) & 650.1 \\
 PGIM & GPT & 0.92 \\
 SCANNER (Ours) & Wiki (Text) & \textbf{1,300.09} \\
\bottomrule
\end{tabular}
}
\caption{Throughput of knowledge extractor. Performance metrics for MoRe are sourced directly from the MoRe paper, while those for PGIM and SCANNER are obtained from our measurements on a Ryzen 7900 CPU. Bigger is faster. For each 100 sentences, PGIM paid \$0.2 for using ChatGPT API.}
\label{Speed_comprasion}
\end{table}
\begin{table}[t]
\centering
\vspace{0.25cm}
\resizebox{0.7\linewidth}{!}{%
\begin{tabular}{lccc}
\toprule
\multirow{2}{*}{Methods} & \multicolumn{3}{|c}{\textbf{Twitter-GMNER}}   \\ 
& \multicolumn{1}{|c}{Pre.}  & Rec.   & \multicolumn{1}{c}{F1}\\
\midrule
SCANNER & 68.34	& 68.71 & 68.52 \\
\quad - TYT & -0.46 & -0.44 & -0.45  \\
\quad - SOC & -0.14 & -0.16 & -0.15 \\
\quad - WKK & -0.11 & -0.17 & -0.14 \\
\quad - ICK & -1.09 & -1.09 & -1.09 \\
\bottomrule
\end{tabular}%
}

\caption{Ablation studies on GMNER datasets. `-TYT' is without trust your teacher method. `-SOC' is without sorting objects by clip. `-WKK' is without Wikipedia knowledge.`-ICK' is without image caption knowledge.} 
\label{ablation study in GMNER}
\end{table} 
\begin{table}[t]
\centering
\vspace{0.25cm}
\resizebox{0.7\linewidth}{!}{%
\begin{tabular}{lccc}
\toprule
\multirow{2}{*}{Object tokens} & \multicolumn{3}{|c}{\textbf{Twitter-GMNER}}   \\ 
& \multicolumn{1}{|c}{Pre.}  & Rec.   & \multicolumn{1}{c}{F1}\\
\midrule
9 & 67.50 & 67.85 & 67.67 \\
12 & 68.06 & 68.43 & 68.24  \\
15 & 68.22 & 68.55 & 68.38 \\
18 & \textbf{68.34} & \textbf{68.71} & \textbf{68.52} \\
21 & \textbf{68.34} & 68.68 & 68.51 \\
\bottomrule
\end{tabular}%
}

\caption{Ablation studies on the number of object tokens.} 
\label{ablation study in object tokens}
\end{table} 

\label{detail hyper-parameter}
We conducted our experiments with hyper-parameter settings as outlined in the following Table~\ref{parametersetting_stage1} and Table~\ref{parametersetting_stage2}, and we utilize AdamW~\citep{loshchilov2018decoupled} optimizer for all tasks. `max objects' refers to the maximum number of object knowledge inputs. We performed a grid search for the learning rate within the range of \([5 \times 10^{-6}, 1\times 10^{-5}]\). We tested batch sizes of 4, 8, and 16 to determine the optimal value, and we explored weight decay within a range of \([0.01, 2]\).

\section{Detailed dataset statistics}
\label{detail data information}
To demonstrate the superiority of our method for various NER tasks, we conduct experiments on a range of datasets. The overall dataset statistics are shown in Table~\ref{Dataset Statistics}, and each task description is in below.

\noindent\textbf{NER dataset.}
CoNLL2003~\citep{tjong-kim-sang-de-meulder-2003}, a dataset with four named entities: PER, LOC, ORG, and MISC. We follow the standard setting~\citep{peters-etal-2017-sem, yan-etal-2021-unified-generativ, shen-etal-2023-diffusion}: use both the train set and dev set for training and evaluate with the test set 

\noindent\textbf{MNER dataset.}
Twitter-2015~\citep{10.5555/3504035.3504731} and Twitter-2017~\citep{luetal2018visual}; collected from social network service posts. 
Like CoNLL2003, it consists of the same four named entity types. 
We operate a train set for training and hyper-parameter tuning using a dev set and evaluate it with the test set.

\noindent\textbf{GMNER dataset.}
Twitter-GMNER~\citep{yuetal2023grounded}, a dataset collected by extracting some of the data from Twitter-2015 and Twitter-2017, and employ bounding box annotation. We operate same validate strategy as MNER.

\section{Compare with MoRe and PGIM}
\label{Compare-with-MoRe-and-PGIM}
\noindent\textbf{Using LLM knowledge.}
With the advancements in LLM, we conducted experiments using GPT knowledge in SCANNER instead of Wikipedia knowledge. We utilize the same knowledge used in PGIM, and the results are in Table~\ref{Using_gpt_knowledge}. We can observe that while applying GPT knowledge in SCANNER increases time and API-related costs, it also enhances performance. Notably, despite the GPT knowledge being tailored for PGIM, its performance in SCANNER is superior.

\noindent\textbf{Time cost in retrieval knowledge.}
We conduct time cost comparisons using retrieval knowledge. As shown in Table~\ref{Speed_comprasion}, our entity-centric approach for integrating Wikipedia information demonstrates a speed advantage over MoRe's BM25 and K-NN based information retrieval and PGIM's GPT-based knowledge creation. This speed benefit arises from our direct method of identifying entity candidates, which allows for immediate retrieval of relevant Wikipedia articles without any bottlenecks.

\section{Ablation study in GMNER}
We conduct additional ablation experiments on the Twitter-GMNER task to evaluate the effectiveness of the proposed method. Effects of each method are shown in Table~\ref{ablation study in GMNER}, and ablation studies on the number of object tokens are shown in Table~\ref{ablation study in object tokens}. These tables substantiate the efficacy of our proposed methodologies in the context of GMNER and show that it is optimal when the number of object tokens is 18.

\begin{figure*}[t]
    \centering
    \includegraphics[width=2 \columnwidth]{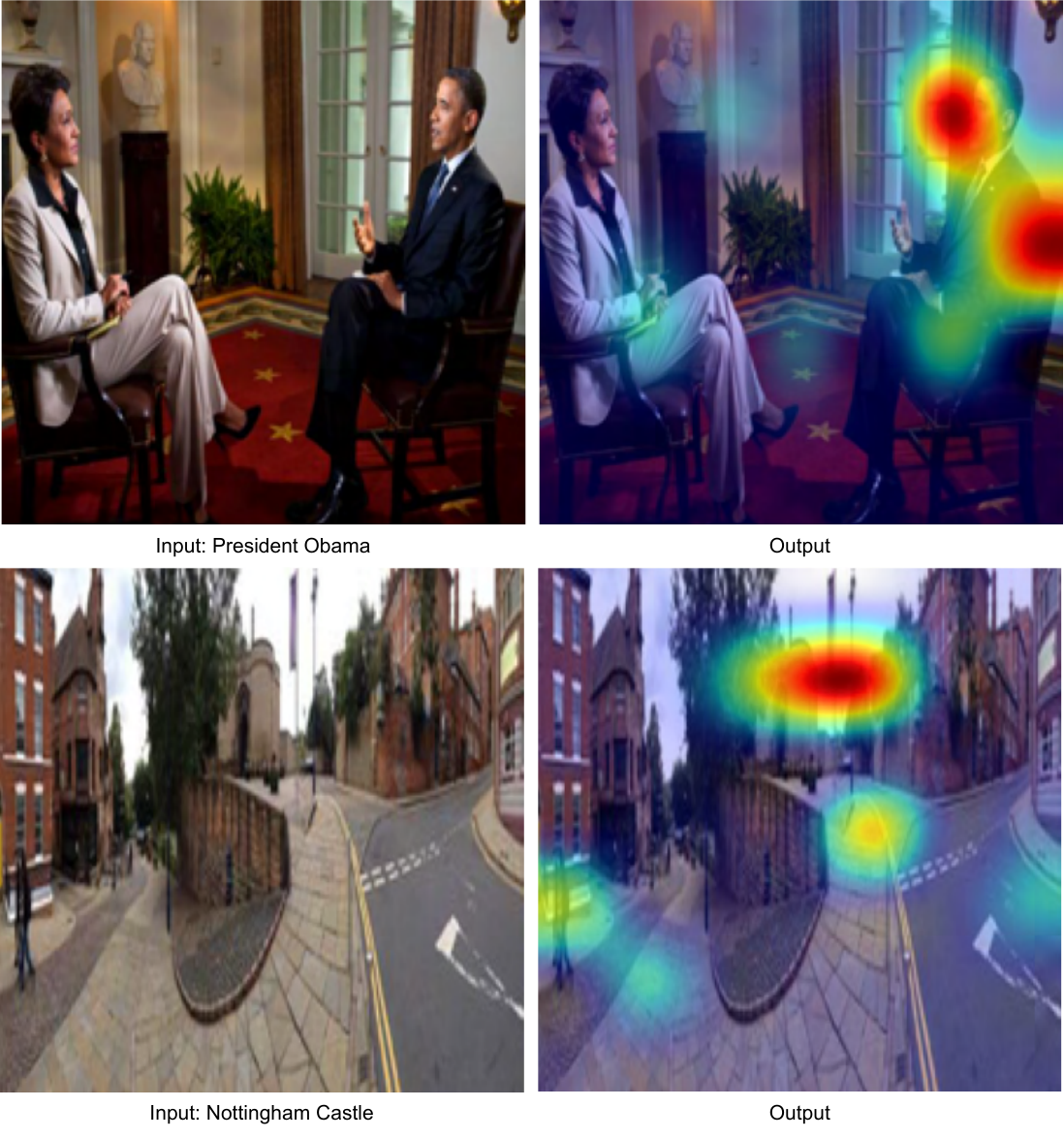}
    \caption{A visualization of which location CLIP focuses on by using Grad-CAM \citep{selvaraju2017_grad}. 
    }
    
    \label{fig:effect_of_clip}
\end{figure*}

\noindent\textbf{Effect of CLIP knowledge.}
CLIP is a practical module extracting entities from an image. As visualized in Fig~\ref{fig:effect_of_clip}, CLIP utilizes its knowledge and attention to the entity location in the image and improves the model's capability in GMNER tasks.

\end{document}